%% file: main.tex
\documentclass[letterpaper]{article}
\usepackage{aaai}
\usepackage{times}
\usepackage{helvet}
\usepackage{courier}
\usepackage{booktabs}
\usepackage{subfiles}
\usepackage{csquotes}
\usepackage[dvipsnames]{xcolor}
\usepackage{graphicx}
\usepackage[thinlines]{easytable}
\usepackage[ruled, linesnumbered]{algorithm2e}
\usepackage{subfigure}
\usepackage{todonotes} 
\usepackage{csquotes}
\usepackage{zref-totpages}
\usepackage{array}
\usepackage{makecell}
\usepackage[dvipsnames]{xcolor}
\newcolumntype{L}[1]{>{\raggedright\let\newline\\\arraybackslash\hspace{0pt}}m{#1}}
\newcolumntype{C}[1]{>{\centering\let\newline\\\arraybackslash\hspace{0pt}}m{#1}}
\newcolumntype{R}[1]{>{\raggedleft\let\newline\\\arraybackslash\hspace{0pt}}m{#1}}

\begin{document}
%

\title{Theme-driven Keyphrase Extraction to Analyze Social Media Discourse} 

\author{William Romano\thanks{Authors contributed equally to the paper.}\textsuperscript{1}, Omar Sharif\textsuperscript{*}\textsuperscript{1}, Madhusudan Basak\textsuperscript{*}\textsuperscript{1,2}, Joseph Gatto\textsuperscript{1}, Sarah Preum\textsuperscript{1} \\
\textsuperscript{1}Department of Computer Science, Dartmouth College, USA \\
\textsuperscript{2}Department of Computer Science and Engineering, BUET, Dhaka, Bangladesh \\
\{william.j.romano.gr,omar.sharif.gr,madhusudan.basak.gr,joseph.m.gatto.gr,sarah.masud.preum\}@dartmouth.edu
}


\maketitle

\subfile{sections/0_Abstract}

\subfile{sections/1_Introduction}

\subfile{sections/2_Related_Works}

\subfile{sections/3_Dataset}

\subfile{sections/5_Experiments}
\subfile{sections/6_Methods}

\subfile{sections/7_Results}
\subfile{sections/8_EthicalConsiderations_BroadImpact}

\subfile{sections/9_Conclusion_and_Future_Work}

\small
\bibliographystyle{aaai}
\bibliography{ref}


\end{document}

%% file: sections/0_Abstract.tex

\begin{abstract}

Social media platforms are vital resources for sharing self-reported health experiences, offering rich data on various health topics. Despite advancements in Natural Language Processing (NLP) enabling large-scale social media data analysis, a gap remains in applying \textit{keyphrase extraction} to health-related content. Keyphrase extraction is used to identify salient concepts in social media discourse without being constrained by predefined entity classes. This paper introduces a theme-driven keyphrase extraction framework tailored for social media, a pioneering approach designed to capture clinically relevant keyphrases from user-generated health texts. Themes are defined as broad categories determined by the objectives of the extraction task. We formulate this novel task of theme-driven keyphrase extraction and demonstrate its potential for efficiently mining social media text for the use case of treatment for opioid use disorder. This paper leverages qualitative and quantitative analysis to demonstrate the feasibility of extracting actionable insights from social media data and efficiently extracting keyphrases using minimally supervised NLP models. Our contributions include the development of a novel data collection and curation framework for theme-driven keyphrase extraction and the creation of \textit{MOUD-Keyphrase}, the first dataset of its kind comprising human-annotated keyphrases from a Reddit community. We also identify the scope of minimally supervised NLP models to extract keyphrases from social media data efficiently. Lastly, we found that a large language model (ChatGPT) outperforms unsupervised keyphrase extraction models, and we evaluate its efficacy in this task.

\end{abstract}

%% file: sections/1_Introduction.tex
\section{Introduction}


Social media platforms like Twitter, Facebook, and Reddit gather spontaneous, self-reported lived experiences from thousands of individuals with a diverse array of medical and socio-demographic conditions. People seeking or undergoing treatment often resort to social media for informational and emotional support \cite{chen2021social}. Findings in public health research reports increased reliance on social media and other online health platforms for addressing health information needs  \cite{Neely2021-ji,Bhandari2014-qg}. Thus social media has become an exceptional source of clinically relevant data to understand population-level concerns, knowledge gaps, treatment perceptions, and barriers \cite{chen2021social}. In addition, social media platforms like Reddit support anonymity and thus encourage rich engagement among peers on stigmatized topics like mental health and substance use recovery \cite{naslund_aschbrenner_marsch_bartels_2016}. Qualitative analysis of social media data has been used to understand various health issues, including COVID-19 \cite{Sleigh2021}, cancer \cite{Levonian_Erikson_Luo_Narayanan_Rubya_Vachher_Terveen_Yarosh_2020}, depression, and other mental health conditions \cite{depression_qual}. 

In parallel, recent advancements in natural language processing have enabled large-scale analysis of social media data, contributing significantly to areas like suicide risk detection, adverse drug reaction detection, and misinformation classification \cite{Mathur_Sawhney_Shah_2020,aroyehun-gelbukh-2019-detection,dharawat2022drink}. However, a significant gap remains in applying keyphrase extraction to self-reported health-related content on social media, including online health communities.

Unlike topic modeling and Medical Named Entity Recognition (MedNER), keyphrase extraction concentrates on identifying salient concepts. This unique property of keyphrases makes it a potentially powerful tool for studying social media discourse, as pre-defined entity classes do not constrain it. Consider a social media post:
\begin{displayquote}
Am I the only one taking \textbf{subs} that feels \textbf{nervus} all of the time? I know \textbf{anxiety} is a symptom of other opioid recovery drugs like \underline{methadone} but I don't take those.
\end{displayquote}
The keyphrases in this post are \textit{subs}, \textit{nervus}, and \textit{anxiety}, where the term \textit{subs} is a shorthand for Suboxone, a medication used to treat opioid use disorder (OUD). Standard MedNER or topic models may struggle to extract these keyphrases if they are not part of other texts in the corpus or are misspelled or presented in shorthand. However, applying keyphrase extraction from social media texts introduces its unique challenges, such as defining what constitutes a \textit{keyphrase}, mitigating annotator biases, and serving end-goal applications.

This paper fills this gap by introducing a novel theme-driven keyphrase extraction framework specifically designed for social media. We present the concept of \textit{themes} as broad categories determined by the objectives of the extraction task. We aim to identify clinically relevant keyphrases that will help uncover knowledge gaps, treatment perceptions, and experiences. This is particularly useful for researchers exploring social media discourse without being confined by pre-defined entity classes. To our knowledge, no existing models currently facilitate theme-driven, efficient keyphrase analysis of social media texts. We exemplify the value of theme-driven keyphrase extraction by analyzing Reddit posts on Medications for Opioid Use Disorder (MOUD), a topic of great relevance in the US due to the escalating opioid crisis.

Our study focuses on the subreddit r/Suboxone, a popular forum for discussing buprenorphine-based prescription medications. Reddit provides an ideal platform for our study due to its anonymous, publicly available relevant data from thousands of affected individuals considering or undergoing buprenorphine-based treatment for opioid use disorder. We employ a mixed-method approach consisting of (i) qualitative exploration of the annotated data to extract clinically relevant insights and (ii) quantitative analysis to demonstrate the feasibility of extracting keyphrases efficiently using minimally supervised NLP models. The quantitative analysis leverages Unsupervised Keyphrase Extraction and Large Language models. We demonstrate the application of Theme-driven Keyphrase Extraction on an online health community and make the following novel contributions:

\begin{enumerate}
    \item We propose a unique framework for data collection and curation for theme-driven keyphrase extraction, encompassing a systematic approach for collecting data from social media, designing a theme-specific schema, annotating the data following the schema, and validating the annotation process.   
    
    \item Leveraging this framework, we develop a dataset, \textit{MOUD-Keyphrase}, comprising human-annotated keyphrases from a Reddit community. This dataset, the first of its kind, brings a new resource to the field and will be made publicly available upon the paper's acceptance.

    \item We demonstrate the effectiveness of theme-driven keyphrases for mining social media data to extract domain-specific insights. Using a qualitative method, this application showcases theme-driven keyphrase extraction and its utility in a real-world context.

    \item We explore the potential of minimal supervision for this task, utilizing ten off-the-shelf Unsupervised Keyphrase Extraction models and ChatGPT. This exploration is a first step towards understanding the role and limitations of minimal supervision in keyphrase extraction tasks.
\end{enumerate}

%% file: sections/2_Related_Works.tex
\section{Related Work}
\subsection{Keyword and Keyphrase Extraction}
Keyphrase extraction has been a widely explored research area \cite{Nomoto2022KeywordEA,hasan-ng-2014-automatic}, with the goal of extracting salient phrases that best summarize a document. Depending on the specific task and how the `keyness' property is defined, the set of extracted keyphrases can vary significantly \cite{firoozeh_nazarenko_alizon_daille_2020}.

Existing keyphrase extraction approaches generally work in two steps. First, they select candidate keyphrases through heuristic rules or manual annotation \cite{wang2016}. Second, they apply supervised \cite{lopez-romary-2010-humb} or unsupervised \cite{ucphrase} approaches to rank keyphrases based on their relevance to the document and return the top-k keyphrases. Studies have also been conducted to expand keyword sets to increase the comprehensiveness of keywords relevant to a specific context \cite{Bozarth_Budak_2022}. Keyphrase extraction techniques can be characterized as statistical \cite{CAMPOS2020257}, graph-based \cite{mihalcea-tarau-2004-textrank}, or embedding-based \cite{bennani-smires-etal-2018-simple} methods. Statistical methods often leverage term frequency and document frequency measures, whereas graph-based methods model words or phrases and their co-occurrence relationships in a graph structure. Embedding-based techniques utilize neural embeddings to capture words' semantic and syntactic properties. Each of these methods has its strengths and weaknesses, and their performance can vary depending on the complexity and context of the text.

Recently, contextual embedding-based approaches have been used for keyphrase extraction \cite{zhang-etal-2022-mderank}. Large language models like ChatGPT have been explored for keyphrase extraction and generation tasks \cite{martínezcruz2023chatgpt}. Early results demonstrate that ChatGPT can achieve state-of-the-art performance on keyphrase generation tasks through simple prompting without additional training or fine-tuning \cite{song2023chatgpt}. 

\subsection{Discourse Analysis for Opioid Use Disorder (OUD)}
The growth of online health communities has provided a space for individuals to share their experiences, provide support, and engage in discussions on topics such as substance use, addiction, and recovery \cite{10.1145/2858036.2858246}. Online health discourse on substance use refers to posts or discussions on social media platforms about drugs and other related topics such as addiction, harm reduction, treatment options, and recovery in social media platforms \cite{Lavertu2021.04.01.21254815}. For example, \citeauthor{CHEN2022100061} \shortcite{CHEN2022100061} analyzed Reddit discussions about cannabis, alcohol, and opioids to examine the nature of stigma related to these substances. \citeauthor{forum77} \shortcite{forum77} found a positive correlation between forum use and recovery by analyzing online discourse.  \citeauthor{OpioidRecovery} \shortcite{OpioidRecovery} investigated posts from opioid recovery subreddits to uncover potential alternative treatment options for OUD. However, our study differs in the NLP task and scope, exploring a broader range of categories beyond treatment options, including psychophysical effects, medical history, and substance dependency \& recovery.  

Overall, our study is unique in several significant ways. Unlike previous studies focusing on author-assigned keyphrases in scientific literature \cite{augenstein-etal-2017-semeval}, we extract keyphrases from rich, online data, which poses unique challenges due to domain-specific vocabulary and colloquial linguistic style, including shorthand, and slang. Additionally, we use thorough manual annotation to identify theme-specific keyphrases instead of Twitter hashtags \cite{zhang2016keyphrase}. Hashtags are not always ideal candidates for theme-driven keyphrase extraction due to over-reliance on popular hashtags, inherent ambiguity, and limited coverage of relevant information. Finally, we propose a framework for curating keyphrases from social media to enhance the interpretability and applicability of keyphrase extraction. Our analysis of the performance and errors of ChatGPT and other unsupervised models aims to contribute to the ongoing characterization of these models' capabilities. This is important given the increasing prevalence of online communities and the need for sophisticated tools to interpret such data.

%% file: sections/3_Dataset.tex
\section{Data Collection and Curation Framework for Theme-driven Keyphrase Extraction}
\label{dataset_creation}
\begin{figure}[h!]
  \centering
\subfigure{\includegraphics[scale=1]{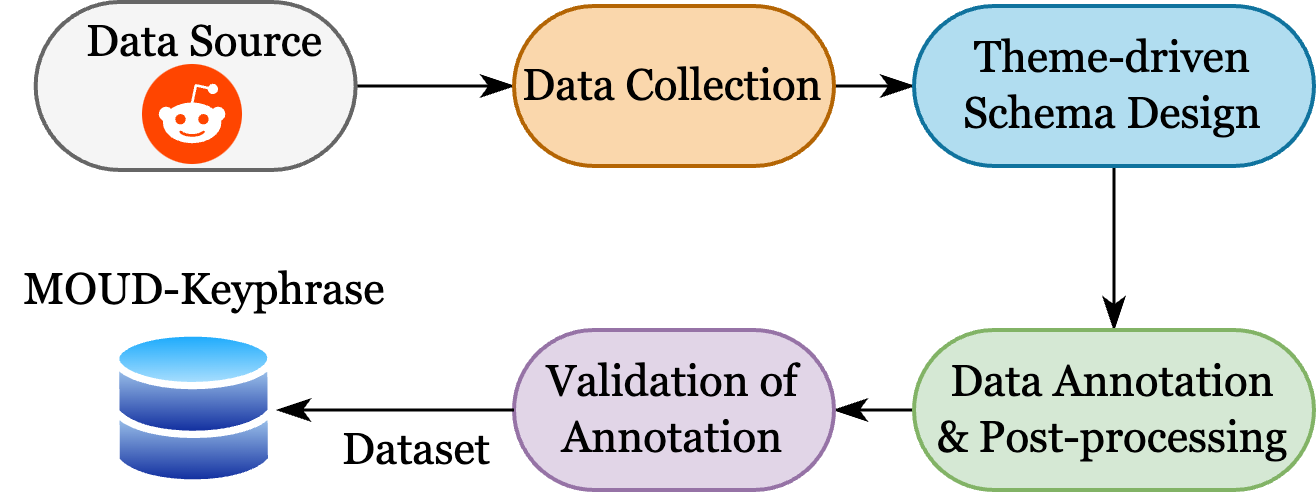}}\quad
\caption{Major steps of theme-driven keyphrase dataset curation framework.} 
 \label{pipeline}
\end{figure}

We develop a new framework for curating theme-driven keyphrase datasets from social media. As shown in Figure \ref{pipeline}, the framework consists of selecting data sources, data collection,  theme-specific schema design, data annotation, and validation of the curated dataset. To determine theme-driven keyphrases, we follow the general criteria for `keyness' provided by \cite{firoozeh_nazarenko_alizon_daille_2020} that include three components `conformity', `homogeneity', and `univocity'.  Conformity is reflected by capturing domain-specific terminology, homogeneity includes normalizing the diverse vocabulary, and univocity refers to specific and non-ambiguous keyphrases.

\subsection{Data Source}
Identifying the sources from where data will be collected is very crucial. It varies on the scope of the problem one is aiming to solve. For example, Twitter can be suitable for collecting brief texts, whereas Reddit is a potential data source for gathering long posts and comments. Data can also be accumulated from multiple sources. Additionally, ethical concerns and biases regarding user data collection must be followed.

To develop our dataset, we choose Reddit since it is anonymous and generates a lot of quality content regarding OUD treatment.
As an anonymous platform, the only public information available for a given user is their username, join date, and activity (i.e., post and comment history). The subreddit from which we collected the data for this study is r/Suboxone, a community with over 30 thousand users as of May 2023. Created in 2011, r/Suboxone is a community forum for dialogue between users who share their relationship with the opioid recovery medication \textit{Suboxone}. This subreddit is strictly moderated, and any posts about buying/selling illicit drugs, exposing other users' personal information, or bullying/abuse are removed. 
Analyzing activity on r/Suboxone makes it possible to achieve a unique and authentic perspective on a user's recovery. While this subreddit explicitly focuses on the treatment of OUD using Suboxone, users frequently discuss their experiences and concerns related to other treatment options too. This is because individuals with OUD may try other OUD treatment options. Discussions extend to various related topics and co-occurring substance and medication usage. 

\subsection{Data Collection}
The subsequent step after selecting the data source is collecting the relevant information. It involves determining the timeline from which data should be curated, identifying features that can facilitate data collection (e.g., flairs, hashtags, and user-assigned tags), and deciding on the specific information to be collected, such as posts, comments, and upvotes. We scraped all posts between the 2nd of January, 2018, and the 6th of August, 2022, on the r/Suboxone subreddit using the PRAW and PushShift APIs \cite{praw_2022,pushshift_2022}.  We observed minimal interaction in this subreddit before 2018 and selected this timeline as it allows us to focus on more recent user data. 
We accumulated posts, comments, likes, upvotes, and unique post ids. Information that violates ethical concerns is not collected or stored. After filtering the corpus for irrelevant posts (e.g., those containing polls, links with no texts, or posts which had been deleted), we developed a corpus of 15,253 posts. Sample posts are presented in Table \ref{table:jaccard-score}.

\subsection{Theme-driven Schema Design}
\label{section:Schema-design}
Determining a phrase to be key to a given text considering various themes is subjective to annotator biases and target downstream applications. It is crucial to design a theme-specific schema to mitigate these biases. According to our exploration, the key steps of the design are: (i) defining the themes with iterative discussion with domain experts, (ii) creating annotation guidelines, and (iii) ensuring the \textit{keyness} of the annotated keyphrases. 

In this study, we aim to extract keyphrases relevant to Suboxone-assisted OUD recovery. We collaborate with a team of five domain experts to define the themes for medication-based OUD recovery. All of our collaborators are well-versed in substance use disorder. Together, they cover a wide variety of expertise, including psychiatry, biomedical data science, digital technology for substance use and mental health, epidemiology, public health policy, addiction medicine, and addiction psychiatry. Two of our collaborators are clinicians and administer MOUD treatment. All of them reviewed our samples and helped us to contextualize different aspects of opioid recovery and the shared lived experiences of the affected individuals. Based on guidance from our collaborators, we identify four main themes: \textit{Treatment Options}, \textit{Substance Dependency \& Recovery}, \textit{Medical History}, and \textit{Psychophysical Effects}. 

The definitions of these themes and the motivation behind choosing them are presented below. Additionally, We have defined another category, \textit{Others}, to capture additional keyphrases that do not belong to any of the four target categories but are still key/salient to the original post. We believe the schema and guidelines presented in this section can be extended to other social communities with health-related discourse.

\begin{itemize}
   \item\textbf{Treatment Options:} This category covers keyphrases related to different treatment options used for recovery. Such as medications used to treat OUD (e.g., \textit{Buprenorphine}, \textit{Methadone}, or their formulations), psycho-therapy, behavioral counseling, or other medications used to cope with withdrawal or other psychophysical effects (e.g., using \textit{melatonin} to help with insomnia while in recovery). We consider prescribed medications, over-the-counter medications, herbal supplements, and other therapeutic options as potential candidates for keyphrases. Keyphrases related to this category can facilitate the analysis of people's perceptions of various treatment options for OUD recovery, including their effectiveness and or lack thereof. 
   
  \item\textbf{Substance Dependency \& Recovery:} This category covers keyphrases related to the history of substance use (e.g., \textit{fentanyl}), co-occurring substance use (e.g., \textit{tobacco}, \textit{alcohol}), and critical factors in recovery (e.g., \textit{relapse}). We consider both prescribed and self-administered substances. Analysis of these keyphrases can aid in identifying salient themes relevant to substance usage history, and trajectory of recovery, e.g. tapering, and relapse.
  
  \item\textbf{Medical History:} Keyphrases concerning medical history, including diagnosis and self-diagnosis of any physical and mental health conditions, relevant medical procedures (e.g., \textit{major surgery}), or critical family medical history. This category covers medical history beyond substance dependency/recovery. Examining medical history-related keyphrases can assist in studying the impact of other health conditions on OUD recovery. 
  
  \item\textbf{Psychophysical Effects:} Keyphrases regarding any physical or psychological effects and symptoms associated with OUD recovery, e.g., psychological effects relevant to withdrawal, precipitated withdrawal, and side effects of medications.  We aim to better understand the common or rare (if any) effects of OUD treatment options by examining these keyphrases. 
  \item\textbf{Others:} All keyphrases related to OUD but do not belong to any of the above four categories are assigned to this category.
\end{itemize}

The insights drawn from analyzing theme-specific keyphrases can aid in devising effective intervention strategies for treatment induction, adherence, and retention. In the second step, we thoroughly iterated over our definition of keyphrases via multiple rounds of trial annotations and discussions between annotators to create the annotation guidelines. Finally, we ensured that keyphrases were annotated to satisfy the general criteria for `\textit{keyness}' inspired by \citeauthor{firoozeh_nazarenko_alizon_daille_2020} \shortcite{firoozeh_nazarenko_alizon_daille_2020}. Conformity, for our purposes here, is reflected by annotating theme-specific terminology. We create homogeneous keyphrases by normalizing slang and misspellings associated with informal discussions online. Finally, for univocity, annotators are instructed to choose non-ambiguous keyphrases. 
 
 

\subsection{Data Annotation and Post-processing} 
Choosing the optimal strategy to annotate theme-specific keyphrases manually can be difficult \cite{firoozeh_nazarenko_alizon_daille_2020}. The most commonly adopted annotation strategy includes annotation by domain experts, hybrid (combination of domain experts and hired annotators), and crowd workers \cite{chau-etal-2020-understanding}. We chose the second option as it yields more high-quality annotation while not putting too much burden on domain experts.

We randomly selected 1,000 posts from the collected samples for keyphrase annotation, a set of comparable size to other notable works in keyphrase extraction \cite{augenstein-etal-2017-semeval}.  Larger social media datasets for keyphrase extraction also exist \cite{zhang2016keyphrase}, but those datasets leverage automated approaches like using \textit{Twitter hashtags} as keyphrases. However, our theme-driven definition of keyphrases requires rigorous manual annotation. Thus we limit our sample size for manual annotation to 1,000 posts. The complete data annotation was manually carried out by four graduate students who are both regular social media users and active in NLP research. They studied/used opioid recovery subreddits, so they possessed better domain knowledge than mTurk annotators. Moreover, they were trained on the annotation task and provided background on MOUD and suboxone through multiple sessions led by experts. We used LightTag, an online platform, to label the keyphrases \cite{LightTag}. Two annotators annotated each sample to generate high-quality keyphrases. Experts resolved any confusion during annotation through discussion. Example keyphrases from each category of MOUD-Keyphrase are exhibited in Table \ref{table:keyphrase-cat-ex}.

\begin{table}[h!]
\centering
\small
\begin{tabular}{L{3cm} L{0.8cm} L{3.5cm}}
\hline
\textbf{Theme}& \textbf{Frequency} & \textbf{Example Keyphrases} \\
\hline                   
Treatment Options & 182 & adderall, antidepressant, naloxone \\
\hline
Substance Dependency \& Recovery & 77 & cocaine, fentanyl, heroin \\
\hline
Medical History & 35 & covid, osteoarthritis, ptsd \\
\hline
Psychophysical Effects & 331 & aches, constipation, panic attack\\
 \hline
 Others & 256 & scam, travel, boyfriend\\
 \hline
\end{tabular}
\caption{Example keyphrases from each theme. Frequency denotes the number of unique keyphrases after normalization for each theme on the dataset.}
\label{table:keyphrase-cat-ex}
\end{table}

\begin{table*}[ht!]
\centering
\footnotesize
\begin{tabular}{L{1.5cm} L{8.7cm} C{2.2cm} C{2.2cm} C{.7cm}}
\hline
\textbf{Title}&\textbf{Post}&\textbf{Annotation-1}&\textbf{Annotation-2}&\textbf{JI}\\
\hline                   
Getting on \textbf{suboxone} & I just came from Florida and have been \textbf{clean} from \textbf{dope} for 9 months but the \textbf{cravings} are setting in. Can anyone suggest me the best way to go about getting on \textbf{suboxone}?? & craving, heroin, clean, suboxone & suboxone, craving, heroin & 0.75\\
\hline
\textbf{Tapering} from 24mg & Will I go through any \textbf{withdrawals} tapering off 24mg of \textbf{suboxone} if I taper down to 22mg? Been on 24mg for 4 months. & suboxone, taper, withdrawal & suboxone, taper, withdrawal &1.0 \\
\hline
\end{tabular}
\caption{\label{result:2} Sample excerpts with titles, posts, and annotations. Annotated keyphrases are normalized (e.g., \textit{dope} is normalized to \textit{heroin}). The Jaccard Index (JI) is calculated over the normalized keyphrase list and indicates the similarity between the annotations.}
\label{table:jaccard-score}
\end{table*}
Depending on the task, post-processing might also be required to ensure the `homogeneity' property of the annotated keyphrases. For our task, we have to normalize the extracted keyphrases manually. Reddit users commonly use various unique phrasings of the same word, e.g., shorthand, slang, and misspellings. For example, the opiate \textit{heroin} may also be referred to as \textit{h}, \textit{dope}, \textit{smack}, and \textit{speedball}, and also it is often misspelled as \textit{herion}, \textit{heroine} etc. 
We manually map all keyphrase variations to their most meaningful representative parent phrases to solve this problem. Due to space constraints, the steps and guidelines we followed to normalize keyphrases are provided in the supplementary materials.


\subsection{Validation of Annotation}
The validation process can include quantitative and qualitative measures to ensure the quality of annotations. Frequently used quantitative measures are inter-annotator agreements, recall, diversity, etc. The qualitative measure depends on the overarching goal and how the keyphrases can be utilized to achieve this. As such, we aim to discover data-driven knowledge from extracted keyphrases, which we demonstrate further through exploratory analysis (RQ1).

Since our annotation involved multiple annotators, we calculated the inter-annotator agreement score to ensure the dataset's quality. We used two lists of normalized keyphrases for each sample from the annotators. We use the Jaccard index to measure the agreement/similarity between annotations \cite{Sarwar2022EvaluatingKE}. Jaccard index is defined as:
\begin{equation}
    JI=\frac{len(A\cap B)}{len(A)+len(B)-len(A\cup B)}  
\end{equation}
Let $A$ and $B$ be the respective set of keyphrases from annotators 1 and 2 for a given sample $i$ in the dataset. $JI$ computes the Jaccard index for sample $i$ and represents the annotator agreement for that sample. While calculating the intersection and union of the two sets, we considered the exact string match between the elements of the sets as used in \citeauthor{Schopf_2022} \shortcite{Schopf_2022}. We used  $Avg.(JI)$ to capture the average Jaccard index for the whole dataset of $ n$ samples. The average Jaccard similarity index obtained using the exact string match approach was $61.36\%$. Given the complexity and subjective judgment of the task, this score indicates a moderate lexical similarity between keyphrases extracted by multiple annotators \cite{Sarwar2022EvaluatingKE}. Sample posts with extracted keyphrases and $JI$-score are presented in Table \ref{table:jaccard-score}.

%% file: sections/6_Methods.tex
\section{Methods}
In this section, we outline our research questions and describe the methodology we employed to address them, utilizing our meticulously compiled \textit{MOUD-keyphrase} dataset. Employing a mixed-method approach, we initially adopt a qualitative strategy to uncover the clinical insights with our first research question. Subsequently, we proceed with a quantitative analysis to address the second research question, examining the potential of minimally supervised NLP models in efficiently extracting keyphrases.

\subsection{RQ1: What clinical insights can theme-driven keyphrases from social media provide?}

 To answer this question, we take a qualitative approach. We look into the list of normalized keyphrases and their frequencies across the dataset for each theme. We investigate frequently-recurring keyphrases as well as those with limited occurrences to gain better insights into the recovery process of OUD. Frequent keyphrases (e.g., \textit{heroin}) offer a ground to study the associated concerns and problems. In contrast, infrequent keyphrases (e.g., \textit{sex drive}) can aid in finding rare, new, and clinically undocumented evidences. Furthermore, we consider the keyphrases that reflect individuals' opinions and emotions toward specific treatment options. We map the keyphrases to the motivating usecases for each theme (referring to the section defining theme-driven schema design) and manually review the associated posts to uncover insightful information about the themes.
 
We also examine the co-occurrence of keyphrases both within and across different themes. This analysis could lead to discovering previously unknown side effects and new treatment options that can be further explored for hypothesis-driven research. 





\subsection{RQ2: Can we effectively extract keyphrases using minimal supervision?}
Qualitative analysis can potentially uncover valuable insights from data, and a comprehensive dataset can aid in grounding these insights. Nonetheless, manual annotation requires a lot of time and effort. Exploring quantitative methods to facilitate data annotation with minimal supervision is crucial. Therefore, we study the scope of unsupervised keyphrase extraction techniques and large language models (e.g., ChatGPT). ChatGPT has demonstrated outstanding performance in NLP tasks across diverse domains \cite{qin2023chatgpt}. However, limited research has evaluated ChatGPT's keyphrase extraction ability on theme-specific long documents from online health communities. We aim to assess the performance of ChatGPT and ten off-the-shelf unsupervised keyphrase extraction models on our dataset. 

\subsubsection{Keyphrase extraction using unsupervised off-the-shelf methods:} We experimented with three types of methods: statistical, graph-based, and embedding. These techniques represent standard approaches in keyphrase extraction and provide a diverse range of technical implementations for thoroughness. We evaluate the performance of two statistical methods, TfIdf \cite{tfidf}, and YAKE \cite{CAMPOS2020257}. We explore four graph-based approaches to keyphrase extraction, namely TextRank \cite{mihalcea-tarau-2004-textrank}, TopicRank \cite{TopicRank}, PositionRank \cite{positionrank}, and MultipartiteRank \cite{DBLP:journals/corr/abs-1803-08721}. These classic models are commonly used as baselines in modern unsupervised keyphrase extraction studies  \cite{zhang-etal-2022-mderank}. We employ the following pipeline for all four embedding methods: 1) Extract candidate keyphrases, 2) Embed each candidate and the full Reddit post, and 3) Return the top-k candidate keyphrases with the highest cosine similarity to the embedding of the entire Reddit post. We evaluate the performance of DistilBERT \cite{distilbert}, BERT \cite{BERT}, DeBERTa-v3 \cite{deberta}, and SBERT \cite{sbert}, as they are all widely-used transformer models with proven ability to produce meaningful contextual word embeddings. 

\subsubsection{Keyphrase extraction using ChatGPT:} We use ChatGPT \cite{ouyang2022training} (OpenAI API model version \textit{gpt-3.5-turbo-0301}) to extract theme-specific keyphrases from the posts. Our experimentation involved both zero-shot and few-shot approaches, using various prompts \cite{brown2020language}. The optimal prompt was determined by a process of trial and error, guided by empirical observations of the model's outputs. We tested several styles of prompts, discarding those that produced outcomes significantly deviating from our expectations and refining those that showed promise. Two different prompt templates, namely \textit{`basic'} and \textit{`guided,'} were utilized to conduct the experiments. In the \textit{`basic'} template, the model prompted to generate OUD-related keyphrases, while in the \textit{`guided'} version, the detailed definitions of various themes were provided in the prompt. The full details and prompt templates are provided with the supplementary materials. The temperature value was set to 0.0 during the experiments, ensuring the deterministic extraction of keyphrases by the model. 

\subsubsection{Experimental Setup:} For the statistical and graph-based methods, we use the PKE library \cite{pke} to perform keyphrase extraction. For the embedding methods, we use KeyBERT \cite{grootendorst2020keybert} to extract and rank keyphrases. All models use the same candidate keyphrases. Specifically, we use PKE's grammar selection tool to extract only noun phrases as candidates for each post. 
Finally, we investigate the performance of ChatGPT in zero-shot and few-shot settings and compare it with other models. 

\subsubsection{Evaluation:} We compare the models using various metrics (i.e., precision, recall, F1 score), but the primary focus will be on the F1 score. This metric considers both precision and recall, providing a more comprehensive view of model performance. The ground truth for these metrics is established based on the human-annotated data. We present the F1@k metric for unsupervised keyphrase extraction models, where the value of \textit{k} represents the top \textit{k} keyphrases extracted by the model. This approach allows us to effectively compare the quality and relevance of the keyphrases produced by each model. In the evaluation of ChatGPT, we assess precision (P), recall (R), and F1 score, as the model does not produce a ranked list of keyphrases.  

By focusing on the effectiveness of different keyphrase extraction models and investigating the conditions under which they excel, we aim to contribute a nuanced understanding of keyphrase extraction from online health communities.

%% file: sections/7_Results.tex
\section{Results}

\subsection{RQ1: What clinical insights can theme-driven keyphrases from social media provide?}

After normalization, we obtained a total of 881 unique keyphrases. The distribution of the keyphrases among different groups can be found in Table \ref{table:keyphrase-cat-ex}. The findings of our thematic analysis of frequent keyphrases are described in the following.

\begin{enumerate}
    \item \textit{Treatment options:} Many users posted about treatment options (182 times), indicating that they were concerned about their treatments and thus tried different treatment options. Commonly observed keyphrases from  this theme were \textit{suboxone}, \textit{kratom}, \textit{buprenorphine}, etc. An interesting observation is that users discussed \textit{subutex} (84) more than \textit{kratom} (69). But using theme-based analysis, we found \textit{suboxone} co-occurred more with \textit{kratom} (61) than \textit{subutex} (57), potentially indicating \textit{kratom} is a more popular alternative of \textit{suboxone}.
    
    \item \textit{Substance dependency \& recovery: }  We also found reasonable evidence (77) of keyphrases related to \textit{substance dependency \& recovery}. Frequently occurring keyphrases in this category are \textit{taper}, \textit{heroin}, \textit{oxycodone}, and \textit{fentanyl}. From the data, we observed that the majority of posts discussed relapsing to \textit{heroin}.
    
    
    \item \textit{Medical history:} We found the lowest number (35) of keyphrases in the \textit{medical history} category. The possible reason is that users focused on the ongoing problems or effects and did not mention medical history, or the reported medical histories were not often the keyphrase in a post. Examples of keyphrases from \textit{medical history} theme include but are not limited to \textit{pregnancy}, \textit{adhd}, and \textit{bipolar disorder}.
    
    \item \textit{Psychophysical effects:} We observed that the highest frequency keyphrase category is \textit{psychophysical effects} with 331 in total. The prevalence of \textit{psychophysical effects} can be attributed to the fact that users discuss different types of effects they had faced due to the concurrent treatment or substance use, thus seeking suggestions. \textit{Withdrawal}, \textit{sleep}, and \textit{precipitated withdrawals} were the top mentioned keyphrases from  this theme. 
    Furthermore, the top keyphrases describing the psychophysical effects of withdrawal are \textit{sleep}, \textit{depression}, and \textit{anxiety}. 
    %
\end{enumerate}


Our systematic theme-specific analysis of frequent keyphrases is valuable for clinicians and researchers focusing on MOUD treatment. For example, it is advantageous for researchers to understand which other treatment options patients consider while undergoing suboxone-based treatment for opioid recovery. This information can uncover potentially harmful trends in self-prescribed or new treatment options critical to a patient's recovery experience. Similarly, quantifying the prevalence of various psychophysical effects can guide researchers toward emerging, potentially significant side effects which require attention from public health officials. Identifying medical/substance use histories can inform about clinically relevant subpopulations seeking MOUD treatment who frequently engage on Reddit, e.g., individuals with fentanyl or heroin dependency, pregnant women seeking MOUD treatment, or individuals interested in tapering their medications. These findings can inform tailored intervention design for MOUD treatment, i.e., who can be reached through Reddit and when.

\noindent
\subsubsection{What insights can we draw from keyphrase co-occurrences?}
Our analysis found that at least one instance of the keyphrase \textit{suboxone} co-occurs with each of the known adult side-effects listed on the U.S. Substance Abuse and Mental Health Service Administration webpage \cite{Samhsa_2022}. This relationship to standard substance abuse guidelines validates the quality of our annotated data. Additionally, such annotations allow one to gather many self-reports based on suboxone and a corresponding psychophysical effect. This facilitates analysis of the context and severity of various opioid-related issues. For example, one user posted: 

\begin{displayquote}
Is anyone getting \textbf{heart or chest pain}, \textbf{back pain} while \textbf{breathing} in sometimes and pain in right upper abdomen randomly while you're \textbf{detoxing} yourself with \textbf{subs}?...day 5 [and] still finding quite a few of these scary symptoms ... 1st time I was kicking \textbf{tranq dope} so idk if it’s just that? or the \textbf{subs} somehow?
\end{displayquote}

Here we find a user experiencing a physical side-effect of starting suboxone (i.e., lingering withdrawal symptoms) being perceived as a general suboxone side-effect. MOUD researchers are interested in such cases as misperceptions in MOUD treatment can negatively impact treatment induction, adherence, and retention. 

Additionally, we find many side effects co-occurring with suboxone which are \textit{not} officially listed as known psychophysical responses to suboxone. For example, almost 2\% of user posts in our dataset expressed issues with their sex drive. Other emerging co-occurring psychophysical effects of suboxone include depression, depersonalization, anger, skin crawling, mood swings, hunger, and memory issues. Researchers interested in monitoring rare adverse drug reactions or unknown drug effects may benefit from keyphrase extraction co-occurrence analysis. Accurate extraction of keyphrases in recovery-related social media discourse can help identify similar discussions and inform the design of qualitative, prospective studies to identify population-level perceptions and misperceptions related to  MOUD treatment.

In addition to co-occurrence patterns with the keyphrase \textit{suboxone}, there are intriguing patterns with \textit{tapering}. Co-occurrence patterns---such as between \textit{tapering} and \textit{withdrawal} are unsurprising given that dose reduction usually induces withdrawal symptoms. More interesting is the co-occurrence pattern that \textit{tapering} has with both \textit{sleep} and \textit{anxiety}. Such significant associations present in the dataset but not explored by the literature are primary candidates for exploration in future work.

\begin{table}[t!]
\small
\centering
\begin{tabular}{@{}lccc@{}}
\toprule
\multicolumn{1}{c}{Model}   & \multicolumn{1}{c}{F1@5} & \multicolumn{1}{c}{F1@10} & \multicolumn{1}{c}{F1@15} \\ \midrule
\textbf{Statistical Methods}&                           &                              &      \\ \midrule
TfIdf  & 0.362 &	0.352 & 0.333 \\     
YAKE  & 0.293	& 0.309	& 0.303\\
\\
\textbf{Graph-Based Methods}&                           &                              &                              \\ \midrule
TextRank & 0.148	& 0.224 & 	0.259     \\
TopicRank & 0.293 &	0.3099 &	0.303 \\
PositionRank     & 0.265 &	0.288	& 0.298         \\
MultipartiteRank & 0.301 & 0.317	& 0.311\\
\\
\textbf{Embedding Methods}  &                           &                              &                             \\ \midrule
DistilBERT  & 0.329 &	0.332 & 0.319 \\
BERT & 0.226 & 0.284 & 0.307 \\
DeBERTa-v3 & 0.207 & 0.249	& 0.281 \\                    
SBERT & 0.250 & 0.319 & 0.332\\
\bottomrule
\end{tabular}
\caption{F1@k for each unsupervised keyphrase extraction model. We varied the value of \textit{k} during testing to evaluate the model's performance using different sets of ranked candidate keyphrases.}
\label{table:UKE}
\end{table}

\subsection{RQ2: Can we efficiently extract keyphrases using minimal supervision?}


Our experimental results exhibit that the TfIdf method produces the highest F1-score across all unsupervised baseline experiments presented in Table \ref{table:UKE}. 
The method that achieved the second-best performance is SBERT. This is expected as 1) embedding approaches are unique in producing contextual representations of the whole post, which explicitly encode textual semantics---a valuable feature for unsupervised keyphrase extraction; 2) SBERT is the only embedding method employed that was trained to output embeddings for longer sequences of texts (other methods are only explicitly trained to produce only high-quality token embeddings). Unfortunately, none of the standard unsupervised models achieved F1 above 0.36. This limits the use of off-the-shelf models to extract theme-driven keyphrases. 

This motivates us to explore the capabilities of large language models (e.g., ChatGPT) for the keyphrase extraction task. The performance of ChatGPT with both zero-short and few-shot approaches is presented in table \ref{GPT-table}. From the results, it is evident that the few-shot examples improve the model's performance. In both few-shot experiments, we randomly sampled three examples from the dataset. The average score of 5 runs with the same temperature is reported to mitigate potential bias to specific samples. The basic prompt with the few-shot examples acquired the highest F1 score of 0.51 and outdid all the other models. The guided few-shot model score is surprisingly low, only 0.271. This could be due to the ChatGPT model being too sensitive to their input prompts. Furthermore, the success of prompting heavily relies on the familiarity of the label space \cite{min-etal-2022-rethinking}. Regarding keyphrase extraction, the label space is vast and challenging to cover with few-shot samples. As a consequence, this might lead to suboptimal performance. 

\begin{table}[t!]
\small
\centering
\begin{tabular}{L{2.5cm}cccc }
\toprule
\textbf{Model}   & \textbf{P} & \textbf{R} & \textbf{F1} \\ \midrule
TfIdf & 0.238 &	0.555 & 0.333 \\
MultipartiteRank & 0.222 &	0.519 &	0.311\\
SBERT &  0.237 &	0.557 &	0.332 \\
\midrule
Zero-shot (Basic) &0.180 &  0.313 & 0.229 \\
Few-shot (Basic) & \textbf{0.478} & 0.554 & \textbf{0.510} \\
Zero-shot (Guided) & 0.162 & 0.597 & 0.256 \\ 
Few-shot (Guided) & 0.173 & \textbf{0.622} & 0.271 \\
\bottomrule
\end{tabular}
\caption{ChatGPT performance comparison on Zero-shot and Few-shot settings with best unsupervised keyphrase extraction models. Since ChatGPT does not provide a ranked set of keyphrases, we focus on precision (P), recall (R), and F1 scores for evaluation purposes. Experiments labeled \textit{basic} were only prompted for basic keyphrase extraction without additional context. Those labeled \textit{guided} utilized theme-specific guidelines in their prompts.}
\label{GPT-table}
\end{table}

Although the basic prompt with the few-shot performs better than the unsupervised models, the overall performance of this model with an F1-score of 0.51 is poor. To get more insights, we further analyze the outputs generated by this setting of ChatGPT. We found that it suffers from issues like missing important keyphrases, focusing on irrelevant keyphrases, showing weaknesses in filtering out specific keyphrases, and performing poorly in determining the occurrences of the keyphrase for all the themes.

\begin{itemize}
    \item \textbf{ChatGPT often misses important keyphrases:} Being a generalized extractor, ChatGPT often misses identifying keyphrases that carry useful information about the specific context. For example, in the following example, \textit{clean} and \textit{pregnancy} carry valuable information about the context of the post, but those have been missed by ChatGPT.

\begin{displayquote}
    ... Like probably many people on this sub and on suboxone in general I have had a long struggle with my addiction. I have been on Suboxone for 3 years and I think that the last three years have been the most stable years that I’ve had with my addiction since it started (other then during my \textbf{pregnancy}, which was before the Suboxone but I am proud to say that I was \textbf{clean} the whole time)...
    \end{displayquote}

\item \textbf{ChatGPT often overpredicts  keyphrases}: ChatGPT often over-predicts keyphrases. For each sample, this model extracts on average 6.25 keyphrases (with a standard deviation of 1.11) while the average number of keyphrases in our ground truth is 4.87. Following is an example where along with some relevant keyphrases ( \textit{upset stomach}, \textit{withdrawals}, \textit{achey}), ChatGPT extracts a lot of irrelevant keyphrases (\textit{oatmeal}, \textit{yougurt}, \textit{protein bars}, \textit{vitamins}) which can confuse the future models trained on the data and lead to an incorrect training.

\begin{displayquote}
…So day 8, the hardest is still adjusting and dealing with insomnia, as well as off and on \textbf{upset stomach}. Can't eat meat yet, lots of \textbf{oatmeal}, bananas, \textbf{yougurt}, \textbf{protein bars} and \textbf{vitamins}. I went 4 nights of 2-3hrs of sleep. It's a response of fight or flight in early \textbf{withdrawals}, but today I felt a bit better. I do get pretty \textbf{achey} and exhausted doing too much. But I can keep up with some chores. Force myself outside for 15min…
\end{displayquote}

\item \textbf{ChatGPT performs poorly for some keyphrases}: Along with the qualitative observations, we tried to quantitatively identify the blindspots of the ChatGPT for theme-driven keyphrase identification. We aimed to determine the keyphrases ChatGPT has a minimal success ratio. We considered the keyphrases which occurred in more than 3 sample texts. For all the runs, we found ChatGPT failed to identify a high number of  keyphrases more than half of the time, e.g., \textit{fear}, \textit{crap}, \textit{hospital}, \textit{cold turkey}, and \textit{doctor}.

\begin{figure} [ht]
\centering
\includegraphics[width=0.42\textwidth]{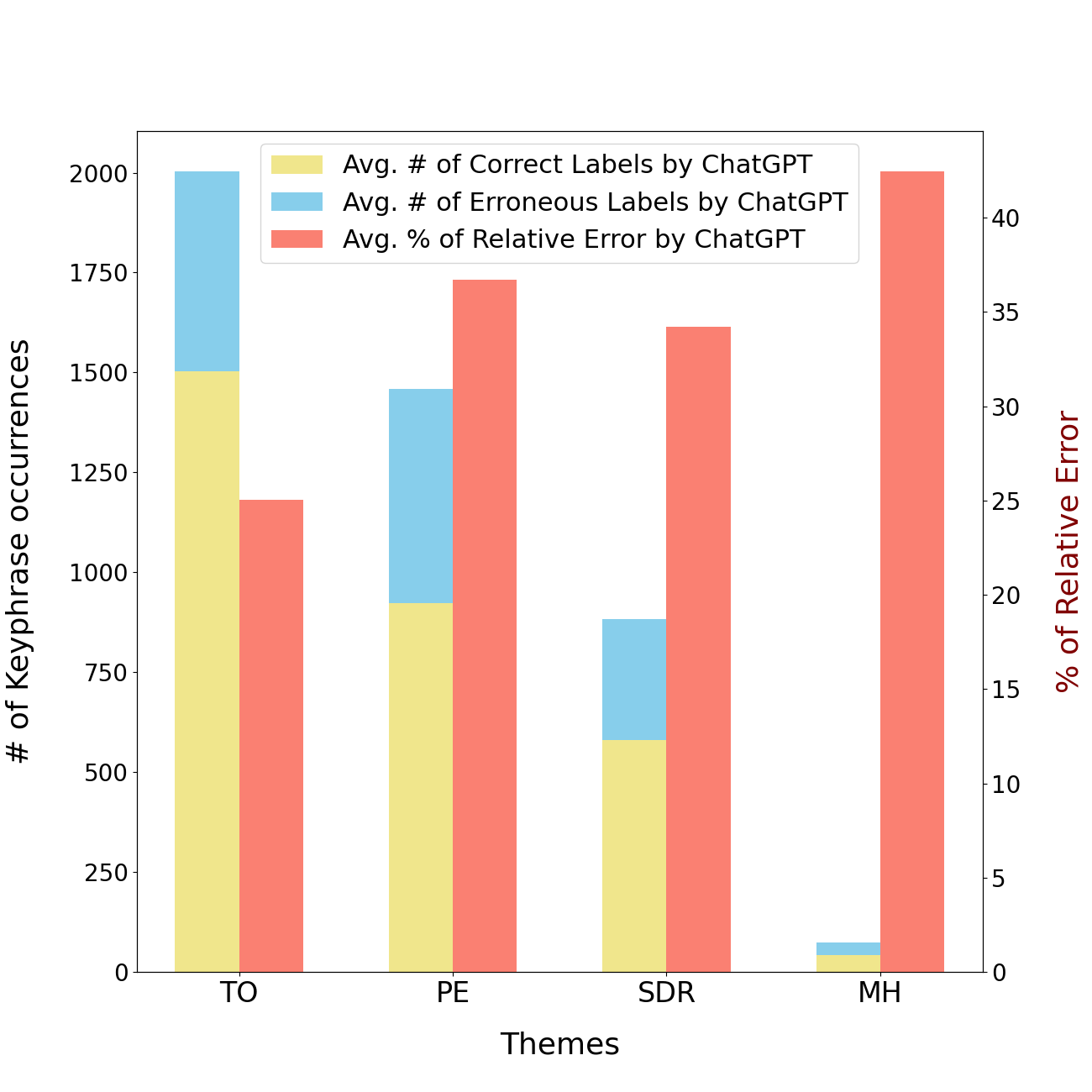}
\caption{The performance of ChatGPT Few-shot (Basic) model in extracting keyphrase occurrence for different themes (TO: Treatment Options, PE: Psychophysical Effects, SDR: Substance Dependency \& Recovery, and MH: Medical History). For each theme, the first bar presents the number of keyphrase occurrences the model failed to identify stacked on the number of keyphrase occurrences correctly identified by it, and the second bar shows the relative error (in percentage) made by the model.
}
\label{fig:chatGPT-error-stat}
\end{figure}

\item \textbf{ChatGPT struggles to determine keyphrase occurrences for all themes:} We were curious to see whether ChatGPT over-predicts or under-predicts any particular themes in our used dataset. Dividing the ground-truth keyphrases of the whole dataset into different themes, we determined the number and percentage of keyphrase occurrences missed by ChatGPT, as shown in Figure \ref{fig:chatGPT-error-stat}. ChatGPT faces more or less difficulties in extracting keyphrase occurrences across all themes. In particular, the model fails to identify a significant percentage of occurrences under the themes of ‘medical history’ (53.15\%) and ‘psychophysical effects’ (41.91\%), highlighting areas for improvement.

\end{itemize}

%% file: sections/8_EthicalConsiderations_BroadImpact.tex
\section{Ethical Considerations}
This research was conducted under Institutional Review Board (IRB) approval at the authors' institution. Since Reddit users can create an account using only an email address, which Reddit does not disclose, the users are largely anonymous. These users can only be identified by the data they voluntarily disclose in their posts. Even if a user shares self-identifying information (e.g., location, age, gender identity), it remains nearly impossible to ascertain the true identity of a Reddit user without further details. This inherent anonymity provides an additional layer of privacy protection.

Regarding the use of Reddit data, the user agreement requires users to consent to share their public posts and comments via the Reddit API, allowing for their utilization in research like ours under specific conditions that respect user privacy and ethical research guidelines. 
Additionally, as a standard precaution, posts included as examples in the paper were paraphrased to prevent the reader from directly identifying the Reddit user account that posted each example.

\subsubsection{Bias and Fairness}

Bias and fairness, especially concerning large models, are increasingly gaining traction in machine learning research \cite{gehman-etal-2020-realtoxicityprompts}. This study bears the potential for bias from three primary sources: research question bias, data bias, and model bias. Research question bias might inadvertently favor conventional clinical relevance as a group of clinical researchers suggested the themes. Data bias can emerge if the training data poorly represents the intended population, for example, data collection focuses only on specific demographics, location, or time that was not intended or part of the study design. With Reddit data typically skewing towards young white males, it could limit its relevance to other demographics \cite{Statista_gender_2022,Statista_age_2021}. Since Reddit data is anonymous, identifying potential bias towards a specific subpopulation is challenging. Model bias could arise if the comparison model is already biased. The absence of control or transparency over the large text corpus used for training some of the large language models could introduce undetected biases, an ongoing issue with large models \cite{dodge-etal-2021-documenting}.


Due to the unavailability of demographic data from Reddit, a complete fairness analysis is currently unfeasible and will be addressed in future research. In response to potential biases, we applied several strategies, including selecting multiple themes, randomly sampling from the more extensive data collection, and diversifying data temporally to minimize time-related bias. Despite these efforts, achieving absolute bias-free research is elusive. We are dedicated to ongoing bias monitoring and mitigation in our studies, hoping that our work aids in detecting and rectifying potential biases in computing research.

\section{Broader Impact}

\paragraph{Implications for keyphrase extraction on Social Media:} Existing work on keyphrase extraction in social media has two limitations: (1) Posts are often derived from Twitter, where low character limits preclude capturing rich information from long, complex self-narrated text from people with lived experiences. Analysis of our data requires understanding texts of variable length, some of which reach even 10,000 characters. Thus it will promote the creation of better HealthNLP models capable of modeling keyphrases in more extended social media discourse on health. (2) Large Twitter datasets utilize hashtags as surrogate keywords---a strategy based on the error-prone assumption that hashtags are always indicators of keyness or saliency. This is the first dataset with keyphrases extracted by human annotators. This work thus provides reliable annotation for clinically relevant keyphrase extraction from Reddit on MOUD-based treatment for opioid recovery.

\noindent
\textbf{Implications for MOUD Research:}
Our \textit{MOUD-Keyphrase} dataset can inform the development of clinician-facing tools that facilitate the discovery of tangible insights that can inform MOUD research and practice. For example, a keyphrase extraction tool based on our dataset could facilitate the discovery of the perceived effectiveness of different MOUD treatment options, strategies to cope with side effects, rare/new adverse drug reactions, and uncover patterns in the patient-reported experience with different MOUDs. Such findings may guide future research in opioid recovery by clinicians and public health researchers. Also, results from such theme-driven keyphrase extraction may guide the development of tailored patient communication tools and programs, e.g., when and how to taper or potentially severe side effects of suboxone. 



%% file: sections/9_Conclusion_and_Future_Work.tex
\section{Limitations and Future Work}

The limitations of this study primarily stem from the inadequate supply of labeled data, which restricts the application of supervised models and might affect the generalizability of our findings. Another limitation is the focus on Reddit posts, potentially missing nuances of health-related discussions on other social media platforms or online health communities. Furthermore, the unsupervised models explored for keyphrase extraction might struggle with Reddit's informal language and prevalent health-related terminology.

Future work could explore the scope of the study to include various social media platforms or online communities to capture a broader spectrum of online health discussions. Annotating more data to enable the deployment of supervised models could also enhance the model's performance. Additionally, developing or adapting keyphrase extraction models to better handle the unique characteristics of social media discourse is a necessary avenue for future research. 

Our study focused on four themes, suggesting that other themes and subthemes in online discourse may have been overlooked. The inconsistent inflections or word choices of extracted keyphrases present another area of improvement. Future studies could leverage large language models for keyphrase generation, and conduct user studies to evaluate the perceived usefulness of extracted keyphrases, thereby enhancing both the quality and utility of the analysis.